\documentclass{article}
\usepackage[tight,footnotesize]{subfigure}
\usepackage{spconf,amsmath,graphicx,amssymb,bm, epsfig}
\usepackage{adjustbox}
\usepackage{scalefnt}
\usepackage{makecell}
\usepackage{epstopdf}
\usepackage{indentfirst}
\usepackage[flushleft]{threeparttable}
\usepackage{float}
\usepackage{algorithm,algcompatible,amsmath,float}
\usepackage{algpseudocode}
\algnewcommand\INPUT{\item[\textbf{Input:}]}%
\algnewcommand\OUTPUT{\item[\textbf{Output:}]}%
\algnewcommand\RETURN{\item[\textbf{Return:}]}%
\algnewcommand\INITIALIZE{\item[\textbf{Initialize:}]}%



\title{Semantic-Preserving Augmentation for Robust Image-Text Retrieval}
\name{Sunwoo Kim, Kyuhong Shim, Luong Trung Nguyen, and Byonghyo Shim}
\address{Department of Electrical and Computer Engineering, Seoul National University, Seoul, Korea\\
\small{\texttt{\{swkim, khshim, ltnguyen, bshim\}@islab.snu.ac.kr}}}

\begin{document}

\maketitle

\begin{abstract}
Image-text retrieval is a task to search for the proper textual descriptions of the visual world and vice versa. One challenge of this task is the vulnerability to input image/text corruptions. Such corruptions are often unobserved during the training, and degrade the retrieval model's decision quality substantially. In this paper, we propose a novel image-text retrieval technique, referred to as robust visual semantic embedding (RVSE), which consists of novel image-based and text-based augmentation techniques called semantic-preserving augmentation for image (SPAug-I) and text (SPAug-T). Since SPAug-I and SPAug-T change the original data in a way that its semantic information is preserved, we enforce the feature extractors to generate semantic-aware embedding vectors regardless of the corruption, improving the model's robustness significantly. From extensive experiments using benchmark datasets, we show that RVSE outperforms conventional retrieval schemes in terms of image-text retrieval performance.
\end{abstract}
\footnote{This work was supported by the National Research Foundation of Korea (NRF) Grant through the Ministry of Science and ICT (MSIT), Korea Government, under Grant 2022R1A5A1027646 and  Future-promising Convergence Technology Pioneer Program of the NRF grant funded by the Korea Ministry of Science and ICT (MSIT) (No. 2022M3C1A3098746).}
\begin{keywords} 
image-text retrieval, data augmentation, robustness, image and text corruption
\end{keywords}

\section{Introduction}
\noindent Recently, image-text retrieval, a task to find out images (sentence) that accurately describe a given sentence (image), has received special attention due to its wide range of applications such as image search, social networking service (SNS) hashtag/post generation, and semantic communication for Internet of Things (IoT), to name just a few~\cite{retrieval:frome2013,retrieval:kli2019, retrieval:semanticcom1,retrieval:semanticcom2, retrieval:semanticcom3}.
Since it is in general very difficult to compare samples obtained from two different modalities (image and text), a projection of image and text to the common embedding space (a.k.a. visual semantic embedding (VSE) space) is required~\cite{retrieval:frome2013,retrieval:kli2019, retrieval:faghri2018, retrieval:wehrmann2019,retrieval:wang2020,retrieval:chen2021}.
To generate the image and text embedding vectors, deep learning (DL)-based image and text feature extractors (i.e., ResNet and BERT) have been popularly used~\cite{retrieval:junhankim2022, retrieval:junhankim2022icassp}.
By comparing the obtained vectors, we can compute the similarity scores between available image-text pairs and then choose the pair with the highest similarity score as the retrieval result.
Main focus of this task is to extract the embedding vectors that can well express a given image and text.

One main challenge in the image-text retrieval is that the DL model is vulnerable to the input image and text corruption.
In many practical scenarios, data corruption occurs frequently. 
For example, there exist a large number of distorted images and unnatural sentences caused by various environmental factors (e.g., illumination, camera motion, weather, and human error).
Even for such corrupted input, one must provide a correct matching answer as long as the semantic meaning of the input does not change. For example, the model should provide accurate image-text retrieval results even if the image is blurry, frayed, or damaged. Unfortunately, if corrupted images and texts are unobserved during the training, these corruptions can degrade the decision process of the image-text retrieval model. This is because the model does not know how to process the input that was not observed during the training, resulting in the generation of a meaningless embedding vector. To evaluate the robustness of various image-text retrieval models to the corruptions, we generate the corrupted datasets by augmenting the original benchmark dataset, Flickr30K~\cite{retrieval:young2014}, with a wide range of image and text corruptions.
Extensive evaluations on
these generated datasets demonstrate that even the latest VSE-based image-text retrieval scheme (e.g., VSE$\infty$ \cite{retrieval:chen2021}) suffers severe performance degradation on corrupted images and texts.

\begin{figure*} [t] 
   \centering
  \epsfig{figure=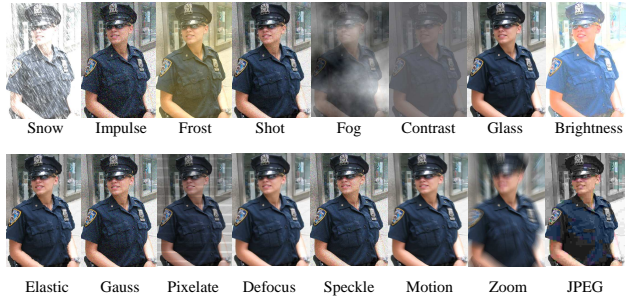, scale=0.25} 
  \vspace{-0.2cm}
  \caption {Examples of image augmentation types (`corrupted-seen') included in the proposed SPAug-I.} 
  \vspace{-0.2cm}
\label{fig:SPAug-I}
\end{figure*}

In this paper, we propose two novel data augmentation techniques, called semantic-preserving augmentation for image (SPAug-I) and text (SPAug-T), to improve the retrieval model's robustness to the image and text corruptions.
Key idea of SPAug-I and SPAug-T is to feed a wide range of semantic-preserving image and text corruptions into the feature extractors as inputs during the training.
By learning from harder training samples, we enforce the feature extractors to generate semantic-aware embedding vectors regardless of the corruption.
Diverse image samples provided by SPAug-I help the feature extractor to focus on the semantic-related information of the original image.
Similarly, semantic-preserving variations of original text data generated by SPAug-T help the text feature extractor to produce the embedding vectors invariant to a wide range of unnatural texts (e.g., sentences with verb tense errors). 
By integrating SPAug-I and SPAug-T to VSE$\infty$, we can develop a more semantic-aware VSE-based image-text retrieval scheme. We henceforth refer it to as robust visual semantic embedding (RVSE).

We mention that the key distinctive feature of our work compared to the conventional augmentations~\cite{ retrieval:devries2017, retrieval:cubuk2018, retrieval:shorten2021, retrieval:wei2019} is that we newly propose the semantic-preserving image and text augmentation techniques tailored for the image-text retrieval.
To be specific, we design SPAug-T such that it changes the given sentence only in a way that its original meaning is preserved. For example, we change the singular/plural forms and verb tenses of the nouns and verbs of the sentence, respectively (e.g.,``My dad catches a fish" $\rightarrow$ ``My dad catches fishes" and ``My dad catches a fish" $\rightarrow$ ``My dad caught a fish", respectively).
Similarly, we design SPAug-I such that it modifies the original image only in a way that can still be described with the same text (e.g., image modifications with impulse, brightness, and zoom noises).
In doing so, we enforce the feature extractors to generate meaningful embedding vectors that capture only the semantic information common across the semantic-preserving variations of the same input.
Since the feature extractors learn to ignore any semantic-irrelevant information that differs in every variation, we can enhance the robustness of retrieval models significantly while improving the retrieval performance on the clean data.


From extensive experiments on two benchmark datasets (MSCOCO-1K~\cite{retrieval:karpathy2015} and Flickr30K~\cite{retrieval:young2014}), we demonstrate that RVSE outperforms the conventional image-text retrieval models.
For example, RVSE achieves about 2\% and 12\% improvements in terms of recall sum (RSUM) over VSE$\infty$, when evaluated on the original and our proposed corrupted Flickr30K test sets, respectively. 
To the best of our knowledge, we are the first to evaluate the performance of image-text retrieval model for corrupted images and texts.

\section{Robust Visual-Semantic Embedding}
\noindent In this section, we present the proposed image-text retrieval scheme called robust visual semantic embedding (RVSE).
We first explain the basics of VSE and then discuss the two major components of RVSE: SPAug-I and SPAug-T.

\begin{table*}[t]
\centering
\label{t3}
\def\arraystretch{1.0} 
\resizebox{1.0\textwidth}{!}{
\begin{tabular}{c|c|c}
\noalign{\smallskip}\noalign{\smallskip}\hline\hline
 Type & Sentence 1 & Sentence 2  \\
\hline
\hline
Original & \makecell{A male is wearing an orange hat and glasses.} & \makecell{A man on a ladder cleans the window of a tall building.} \\
\hline
\makecell{Synonym Replacement} & \makecell{A \textbf{guy} is wearing a \textbf{tangerine cap} and glasses.} & \makecell{A \textbf{male} on a \textbf{ravel} cleans the window of a tall building.}  \\
\hline
\makecell{Article Removal}  & \makecell{Male is wearing orange hat and glasses.} & \makecell{Man on ladder cleans window of tall building.}  \\
\hline
\makecell{Back-Translation}  & \makecell{A \textbf{gentleman wears} an orange hat and glasses.
} & \makecell{A man on a ladder cleans the window of a \textbf{tower block}. }  \\
\hline
\makecell{Be Verb Error}  & \makecell{A male \textbf{am} wearing an orange hat and glasses.
} & \makecell{A man on a ladder cleans the window of a tall building.
}  \\
\hline 
\makecell{Verb Tense Error}  & \makecell{A male \textbf{was} wearing an orange hat and glasses.
} & \makecell{A man on a ladder \textbf{cleaned} the window of a tall building.
}  \\
\hline
\makecell{Singular/Plural Error}  & \makecell{A male wears an orange \textbf{hats} and \textbf{glass}.
} & \makecell{A \textbf{men} on a \textbf{ladders} cleans the \textbf{windows} of a tall \textbf{buildings.}}  \\
\hline
\end{tabular}
}
\caption{Examples of text augmentation types included in the proposed SPAug-T.}
\label{fig:SPAug-T}
\end{table*}

\subsection{Basics of VSE}
\noindent We briefly walk through the formulation of visual semantic embedding (VSE). 
A VSE-based model relies on two feature extractors for each domain. 
First, the image feature extractor $\mathbf{\Phi}$ generates the set of $N$ image features (image grid or object-level features) $\lbrace \phi_{n} \rbrace^{N}_{n = 1}$ from $N$ non-overlapping grids or overlapping boxes $\lbrace \mathbf{i}_{n} \rbrace^{N}_{n = 1}$ of an input image $I$. Second, the text feature extractor $\mathbf{\Psi}$ finds out the set of $M$ text features $\lbrace \psi_{n} \rbrace^{M}_{m = 1}$ from $M$ word tokens $\lbrace \mathbf{t}_{n} \rbrace^{M}_{m = 1}$ of an input sentence $T$.
That is, 
\begin{align}
 \mathbf{\Phi}(I) &= \lbrace \phi_{n} \rbrace^{N}_{n = 1} = \lbrace \mathbf{\Phi}(\mathbf{i}_{n}) \rbrace^{N}_{n = 1}, \label{eq:visualfeatures}\\
 \mathbf{\Psi}(T) &= \lbrace \psi_{m} \rbrace^{M}_{m = 1} = \lbrace \mathbf{\Psi}(\mathbf{t}_{m}) \rbrace^{M}_{m = 1}, \label{eq:textfeatures} 
\end{align}
where $\phi_{n} \in \mathbb{R}^{d_{i}}$, $\psi_{n} \in \mathbb{R}^{d_{t}}$, and $d_{i}$ and $d_{t}$ are the image and text feature dimensions, respectively. 

The extracted features are then aggregated by an image feature aggregator ($f_{\text{VIS}}$) and a text feature aggregator ($f_{\text{TEXT}}$).
These aggregators can be built with average pooling, which computes the mean value among the $N$ ($M$) elements, for each index of the global embedding vector~\cite{retrieval:chen2021}.
Basically, they combine multiple features into two global embedding vectors $e_{i} \in \mathbb{R}^{d_{e}}$ and $e_{t} \in \mathbb{R}^{d_{e}}$, respectively:
\begin{align}
 e_{i} &= f_{\text{VIS}}\big(\lbrace \phi_{n} \rbrace^{N}_{n = 1} \big),  \label{eq:visfinalfeature} \\  
 e_{t} &= f_{\text{TEXT}}\big( \lbrace \psi_{n} \rbrace^{M}_{m = 1} \big), \label{eq:textfinalfeature}
\end{align}
where $d_{e}$ is the dimension of the common embedding space. 
As a feature aggregator, a generalized pooling operator (GPO)~\cite{retrieval:chen2021}, an efficient trainable module that automatically figures out the best pooling function for aggregation, can be employed.

In \eqref{eq:visfinalfeature} and \eqref{eq:textfinalfeature}, one can easily compute the image-text similarity score, which is typically defined as the cosine similarity between $e_{i}$ and $e_{t}$:
\begin{equation}
S( I, T) = \frac{e^{T}_{i}e_{t}}{ \left\lVert e_{i}\right\rVert \cdot \left\lVert e_{t}\right\rVert}.
\label{eq:similarityequation}
\end{equation}
During the inference, the similarity scores are used to rank 1) all candidate images against a query sentence  ($T \rightarrow I$) and 2)  all candidate sentences against a query image ($I \rightarrow T$). 
The top candidates with the highest score are returned as the predicted retrieval result.

\subsection{Semantic-Preserving Image and Text Augmentation}
\noindent In this subsection, we discuss how the proposed SPAug-I and SPAug-T are designed. 
Basically, in SPAug-I, we augment the given original image $I$ of the training data with one of the 16 image noises: 1) Gaussian noise, 2) shot noise, 3) impulse noise, 4) speckle noise, 5) defocus blur, 6) frosted glass blur, 7) motion blur, 8) zoom blur, 9) snow, 10) frost, 11) fog, 12) brightness, 13) contrast, 14) elastic, 15) pixelate, and 16) jpeg compression, each chosen randomly with equal probability (0.5/16 = 0.03125) (see Fig.~\ref{fig:SPAug-I}).
Note that we do not augment the original image data with the remaining probability of 0.5.
To generate diverse images during training, such augmentations are conducted at randomly varying levels of corruption severity as in \cite{retrieval:hendrycks2019}.


In SPAug-T, we augment the original sentence $T$ with one of the 6 semantic-preserving text noises: 1) synonym replacement, 2) article removal, 3) back-translation, 4) be verb error, 5) verb tense error, and 6) singular/plural error, each chosen with equal probability (0.5/6 $\approx$ 0.833) (see Table~\ref{fig:SPAug-T}).
As in SPAug-I, in SPAug-T, we do not augment the original text data with the remaining probability of 0.5.
For synonym replacement and back-translation, we use synonym augmentation and back-translation provided by \texttt{nlpaug} library.
For article removal, we remove all articles in the given sentence.
For be verb error, we swap existing be verbs of the sentence with some other be verb (say, is $\rightarrow$ am).
For verb tense error, we import all verb conjugate forms using the \texttt{pattern.en} module in Python, and then randomly swap all verbs of the sentence with one of their respective conjugate forms.
Lastly, in singular/plural error, we identify all nouns in the sentence and change their original singular/plural forms.

Key benefits of SpAug-I and SPAug-T is that they exclude image manipulations and sentence errors that might degrade the retrieval model's decision quality substantially. 
For example, in SPAug-I, we do not partially mask or randomly crop the given image, as these manipulations can change the image in a way that can no longer be described with the same text.
Similarly, in SPAug-T, we do not change the spelling or randomly swap words of the sentence (e.g., ``My dad catches a fish" $\rightarrow$ ``My dad clenches a fist" and ``My dad catches a fish" $\rightarrow$ ``My fish catches a dad", respectively) since these errors can change what the original sentence meant.



\section{Image-Text Retrieval Experiments}

\subsection{Datasets for Image-Text Retrieval} 
\noindent We evaluate the performance of RVSE using two benchmark datasets: MSCOCO-1K~\cite{retrieval:karpathy2015} and Flickr30K~\cite{retrieval:young2014}.
MSCOCO-1K contains 123,287 images, where each image is paired with 5 sentences.
This dataset is split into 113,287 images for training, 5,000 images for validation, and 5,000 images for testing.
We report results by averaging over 5 folds of 1,000 test images, following the common procedure \cite{retrieval:faghri2018}.
Flickr30K contains 31,700 images, with each one paired with 5 sentences.
Following the data split in~\cite{retrieval:frome2013}, we use 1,000 images for validation and testing, and the rest for training.

To evaluate the retrieval performance of a VSE-based model on corrupted images and texts, we augment the original Flickr30K test dataset, which we refer to as the `clean' dataset.
First, we generate a `corrupted-seen’ image test dataset consisting of 16 test dataset variants that are augmented with 16 types of augmentation included in SPAug-I.
We call this dataset `seen’ because the applied augmentations are observed during the training.
We then generate 6 `corrupted-unseen’ and 6 `corrupted-mixed’ test datasets by augmenting images with the following 12 types of augmentations excluded in SPAug-I: 1) Gaussian blur, 2) spatter, 3) saturate, 4) average blur, 5) median blur, 6) dropout, 7) zoom + snow, 8) zoom + pixelate, 9) zoom + contrast, 10) snow + pixelate, 11) snow + contrast, and 12) pixelate + contrast.
Mixed augmentation types are the composition of two seen image augmentations; we choose zoom, snow, contrast, and pixelate.
Similarly, we come up with 6 text datasets by augmenting sentences with text augmentations of SPAug-T\footnote{To obtain the datasets,  check https://github.com/islab-github/cDatasets.}.

\subsection{Implementation Details of RVSE} \noindent For the choice of image feature extractor $\mathbf{\Phi}$, we test two options: ResNet-101 and ResNet-101 + Faster-RCNN~\cite{retrieval:ren2015}, that correspond to the grid and region image features, respectively.
We apply our method for both options to demonstrate the effectiveness of SPAug-I and SPAug-T regardless of the type of image feature used. 
For ResNet-101, we extract the set of $N$ grid image features~\cite{retrieval:jiang2019} $\lbrace \phi_{n} \rbrace^{N}_{n = 1}$ from $N=64$ non-overlapping grids $\lbrace \mathbf{i}_{n} \rbrace^{N}_{n = 1}$ of an input image $I$.
Similarly, for ResNet-101 + Faster-RCNN, we extract the set of $N=36$ region (object-level) image features $\lbrace \phi_{n} \rbrace^{36}_{n = 1}$ from 36 detected object boxes $\lbrace \mathbf{i}_{n} \rbrace^{36}_{n = 1}$ of an input image $I$.
For the choice of text feature extractor $\mathbf{\Psi}$, we employ BERT~\cite{retrieval:devlin2019}.
In doing so, we extract the set of $M$ text features $\lbrace \psi_{n} \rbrace^{M}_{m = 1}$ from $M$ words $\lbrace \mathbf{t}_{n} \rbrace^{M}_{m = 1}$ of an input sentence $T$.

We evaluate the performance of RVSE using  Recall@K (R@K), which is defined as the proportion of queries whose (any) ground truth is ranked within the top-K retrieved results.
Following the previous works, we use $\text{K} = \lbrace 1,5,10 \rbrace$ for both image to text (Img$\rightarrow$Text) and text to image (Text$\rightarrow$Img) retrievals.
As a summarizing metric to compare retrieval schemes, we use RSUM, the sum of R@1, R@5, and R@10 of both image-to-text and text-to-image retrievals.

\setlength{\tabcolsep}{4pt}
\begin{table}[t]
\begin{center}
\def\arraystretch{1.3}
\begin{adjustbox}{width=\linewidth}
\begin{tabular}{lccccccccc}
\hline
\noalign{\smallskip} 
Method & Image & Text & \multicolumn{3}{c}{Image $\rightarrow $ Text} & \multicolumn{3}{c}{Text $\rightarrow $ Image} & \\
\noalign{\smallskip}
 & Feature & Encoder & R@1 & R@5 & R@10 & R@1 & R@5 & R@10 & RSUM\\
\noalign{\smallskip}
\hline
\noalign{\smallskip}
VSE++$^{\dagger}$ \cite{retrieval:faghri2018} & Grid & BERT & 68.5 & 92.4 & 97.1 & 55.9 & 87.3 & 94.3 & 495.6\\
VSE$\infty^{\dagger}$ \cite{retrieval:chen2021} & Grid & BERT & 70.8 & 93.2 & 97.3 & 58.1 & 88.2 & 95.0 & 502.6\\
\textbf{RVSE (Ours)} & Grid & BERT & \textbf{71.9} & \textbf{94.7} & \textbf{98.8} & \textbf{59.3} & \textbf{89.5} & \textbf{96.3} & \textbf{510.5}\\
\hline
\end{tabular}
\end{adjustbox}
\caption{Performance comparison of VSE-based methods on MSCOCO-1K dataset. $^{\dagger}$ indicates that the results are based on our re-implementation.}
\label{table:maincoco}
\end{center}
\end{table}
\setlength{\tabcolsep}{1.4pt}
\setlength{\tabcolsep}{4pt}
\begin{table}[t]
\begin{center}
\def\arraystretch{1.3}
\begin{adjustbox}{width=\linewidth}
\begin{tabular}{lccccccccc}
\hline
\noalign{\smallskip}
Method & Image & Text & \multicolumn{3}{c}{Image $\rightarrow $ Text} & \multicolumn{3}{c}{Text $\rightarrow $ Image} & \\
\noalign{\smallskip}
 & Feature & Encoder & R@1 & R@5 & R@10 & R@1 & R@5 & R@10 & RSUM\\
\noalign{\smallskip}
\hline
\noalign{\smallskip}
VSE++$^{\dagger}$ \cite{retrieval:faghri2018} & Grid  & BERT & 56.7 & 81.6 & 89.1 & 43.6 & 74.6 & 83.7 & 429.2\\
VSE$\infty^{\dagger}$ \cite{retrieval:chen2021} & Grid & BERT & 66.3 & 88.7 & 93.6 & 50.6  & 78.7 & 86.7 & 464.6 \\
\textbf{RVSE (Ours)} & Grid & BERT & \textbf{68.6} & \textbf{89.7} & \textbf{93.9} & \textbf{51.8} & \textbf{79.9} & \textbf{88.3} & \textbf{472.2}\\
\hline
\noalign{\smallskip}
VSE++ \cite{retrieval:faghri2018} & Region & BiGRU & 62.2 & 86.6 & 92.3 & 45.7 & 73.6 & 81.9 & 442.3\\
LIWE  \cite{retrieval:wehrmann2019} & Region & BiGRU & 69.6 & 90.3 & 95.6 & 51.2 & 80.4 & 87.2 & 474.3\\
VSRN \cite{retrieval:kli2019} & Region & BiGRU & 71.3 & 90.6 & 96.0 & 54.7 & 81.8 & 88.2 & 482.6\\
CVSE \cite{retrieval:wang2020} & Region & BiGRU & 70.5 & 88.0 & 92.7 & 54.7 & 82.2 & 88.6 & 476.7\\
VSE$\infty$ \cite{retrieval:chen2021} & Region & BiGRU & 76.5 & 94.2 & 97.7 & 56.4 & 83.4 & 89.9 & 498.1\\
VSE++ \cite{retrieval:faghri2018} & Region & BERT & 63.4 & 87.2 & 92.7 & 45.6 & 76.4 & 84.4 & 449.7\\
VSE$\infty$ \cite{retrieval:chen2021} & Region & BERT & 81.7 & 95.4 & 97.6 & 61.4  & 85.9 & 91.5 & 513.5\\
\textbf{RVSE (Ours)} & Region & BERT & \textbf{82.5} & \textbf{96.8} & \textbf{99.0} & \textbf{62.6} & \textbf{86.8}  & \textbf{92.9} &\textbf{520.6}\\
\hline
\end{tabular}
\end{adjustbox}
\caption{Performance comparison of non-VSE and VSE-based methods on Flickr30K dataset. $^{\dagger}$ indicates that results are based on our re-implementation.}
\label{table:mainflickr}
\end{center}
\vspace{-0.3cm}
\end{table}
\setlength{\tabcolsep}{1.4pt}
\setlength{\tabcolsep}{4pt}
\begin{table}[t]
\begin{center}
\def\arraystretch{1.3}
\begin{adjustbox}{width=\linewidth}
\begin{tabular}{llccccccc}
\hline
\noalign{\smallskip}
Method & Test Configuration & \multicolumn{3}{c}{Image $\rightarrow $ Text} & \multicolumn{3}{c}{Text $\rightarrow $ Image} & \\
\noalign{\smallskip}
 &  & R@1 & R@5 & R@10 & R@1 & R@5 & R@10 & RSUM \\
\noalign{\smallskip}
\hline
\noalign{\smallskip}
VSE$\infty$  & I(corrupted, seen) - T(clean) & 49.4 & 72.3 & 79.5 & 40.0 & 66.9 & 76.2 & 384.3 \\
 & I(clean) - T(corrupted)  & 62.2 & 86.6 & 92.3 & 46.5  & 74.5 & 83.5 & 445.8 \\
 & I(corrupted, unseen) - T(clean) & 53.5 & 78.6 & 86.2 & 43.5  & 72.1 & 81.2 & 415.1 \\
 & I(corrupted, mixed) - T(clean) & 27.6 & 47.2 & 55.2 & 25.0  & 47.2 & 57.0 & 259.2 \\
\hline
\noalign{\smallskip}
\textbf{RVSE}  & I(corrupted, seen) - T(clean) & 60.2 & 82.8 & 88.4 & 46.2  & 74.5 & 83.4 & \textbf{435.5} \\
 & I(clean) - T(corrupted)  & 67.2 & 89.0 & 93.8 & 50.2  & 78.5 & 87.1 & \textbf{465.8} \\
 & I(corrupted, unseen) - T(clean) & 61.6 & 84.7 & 91.0 & 47.1  & 76.3 & 85.2 & \textbf{445.9} \\
 & I(corrupted, mixed) - T(clean) & 43.4 & 67.9 & 77.3 & 34.7  & 62.1 & 73.1 & \textbf{358.5} \\
 
\hline
\end{tabular}
\end{adjustbox}
\caption{Performance comparison of VSE-based methods on four cases of corrupted Flickr30K test dataset. Here, I(image)-T(text) represents the corruption type of each modality.}
\label{table:mainflickrnoisy}
\end{center}
\vspace{-0.3cm}
\end{table}
\setlength{\tabcolsep}{1.4pt}

\subsection{Comparison of RVSE with State-of-the-art}

\noindent In Table~\ref{table:maincoco}, we  show the performance on the MSCOCO-1K dataset on RVSE and previous VSE-based image-text retrieval techniques: VSE++~\cite{retrieval:faghri2018} and VSE$\infty$~\cite{retrieval:chen2021}\footnote{Since previous works do not report the performance on grid image features, we re-implement VSE++ and VSE$\infty$ as described in \cite{retrieval:chen2021} using the official code of VSE$\infty$.}.
From the results, we observe that the proposed RVSE outperforms the previous VSE-based techniques by a considerable margin on the MSCOCO-1K dataset when using grid image features.
While RVSE exploits the same architecture as VSE$\infty$, the proposed RVSE training strategy has led to performance improvement of about 2\% in terms of RSUM.

In Table~\ref{table:mainflickr}, we compare the performance on the Flickr30K dataset on RVSE and the five VSE-based methods: VSE++~\cite{retrieval:faghri2018}, LIWE~\cite{retrieval:wehrmann2019}, VSRN~\cite{retrieval:kli2019}, CVSE~\cite{retrieval:wang2020}, and VSE$\infty$~\cite{retrieval:chen2021}.
We observe that the RSUM of RVSE is about 2\% higher than that of VSE$\infty$, on both grid and region image features.

To evaluate the generalization capability of RVSE, in Table~\ref{table:mainflickrnoisy}, we compare VSE$\infty$ and RVSE on four types of retrieval tasks: 1) corrupted-seen image-text retrieval, 2) image-corrupted text retrieval, 3) corrupted-unseen image-text retrieval, and 4) corrupted-mixed image-text retrieval (see Section 3.1 for the details).
From the results, we observe that RVSE provides a significant gain over VSE$\infty$ by a large margin.
For example, RVSE achieves around 12\% and 4\% improvement in the RSUM over VSE$\infty$ in corrupted-seen image-text and image-corrupted text retrieval tasks, respectively. 
Even when evaluated on test datasets consisting of images with unseen and mixed augmentations, the RSUM of RVSE is at least 7\% and 28\% higher than that of VSE$\infty$, respectively. 
This implies that RVSE is not over-fitted to certain corruption types and can be well generalized to other corruption types that are not observed during the training.

\section{Conclusion}
\noindent In this paper, we presented a new robust retrieval technique, termed RVSE, which consists of image-based and text-based augmentation techniques tailored for image-text retrieval. 
To provide a correct retrieval result regardless of input corruption, RVSE observes a wide range of semantic-preserving augmentation types during the training.
From the experimental results on benchmark datasets, we demonstrated that our approach improves the model's robustness to the corruptions and outperforms the latest retrieval schemes significantly. 




\begin{thebibliography}{00}

\bibitem{retrieval:frome2013} A. Frome, G. S. Corrado, J. Shlens, S. Bengio, J. Dean, M. Ranzato, and T. Mikolo, ``Devise: A deep visual-semantic embedding model," \emph{Advances in Neural Information Processing Systems}, pp. 2121-2129, 2013.

\bibitem{retrieval:kli2019} K. Li, Y. Zhang, K. Li, Y. Li, and Y. Fu, ``Visual semantic reasoning for image-text matching," in \emph{Proc. ICCV}, pp. 4654-4662, 2019.

\bibitem{retrieval:semanticcom1} S. Kim, L. T. Nguyen, J. Kim, and B. Shim, “Deep learning based low-rank matrix completion for IoT network localization,” \textit{IEEE Wireless Commun. Lett.}, vol. 10, no. 10, pp. 2115-2119, Oct. 2021.

\bibitem{retrieval:semanticcom2} M. Karimzadeh-Farshbafan, W. Saad, and M. Debbah, “Curriculum learning for goal-oriented semantic communications with a common language," \emph{IEEE Trans. Commun.}, Jan. 2023.

\bibitem{retrieval:semanticcom3} S. Kim and B. Shim, “Localization of Internet of Things network via deep neural network based matrix completion,” in \emph{Proc. Int. Conf. Inf. Commun. Technol. Converg. (ICTC)}, Oct. 2020.


\bibitem{retrieval:faghri2018} F. Faghri, J. D. Fleet, R. Kiros, and S. Fidler, ``Vse++: Improving visual-semantic embeddings with hard negatives," in \emph{Proc. BMVC}, 2018.




\bibitem{retrieval:wehrmann2019} J. Wehrmann, D. M. Souza,  M. A. Lopes, and R. C. Barros, ``Language-agnostic visual-semantic embeddings," in \emph{Proc. ICCV}, pp. 5804-5813, 2019.

\bibitem{retrieval:wang2020} H. Wang, Y. Zhang, Z. Ji, Y. Pang, and L. Ma, ``Consensus-aware visual-semantic embedding for image-text matching," in \emph{Proc. ECCV}, pp. 18-34, 2020.

\bibitem{retrieval:chen2021} J. Chen, H. Hu, H. Wu, Y. Jiang, and C. Wang, 
``Learning the best pooling strategy for visual semantic embedding," in \emph{Proc. CVPR}, pp. 15789-15798, 2021.


\bibitem{retrieval:junhankim2022} J. Kim, K. Shim, and B. Shim, 
``Semantic feature extraction for generalized zero-shot learning," in \emph{Proc. AAAI}, pp. 1166-1173, 2022.

\bibitem{retrieval:junhankim2022icassp} J. Kim and B. Shim, 
``Generalized zero-shot learning using conditional wasserstein autoencoder," in \emph{Proc. ICASSP}, pp. 3413-3417, 2022.


\bibitem{retrieval:young2014} P. Young, A. Lai, M. Hodosh, and J. Hockenmaier,
``From image descriptions to visual denotations: New similarity metrics for semantic inference over event descriptions,"
\emph{Transactions of the Association for Computational Lingustics}, vol. 2, pp. 67-78, 2014.

\bibitem{retrieval:devries2017} T. DeVries and G. W. Taylor, ``Improved regularization of convolutional neural networks with cutout," \emph{ arXiv:1708.04552}, 2017.

\bibitem{retrieval:cubuk2018} E. D. Cubuk, B. Zoph, D. Mane, V. Vasudevan, and Q.V. Le, ``Autoaugment: Learning
augmentation policies from data," \emph{arXiv:1805.09501}, 2018.




\bibitem{retrieval:shorten2021} C. Shorten, T. M. Khoshgoftaar, and B. Furht, ``Text data augmentation for deep learning," \emph{Journal of big Data}, vol. 8, no. 1, pp. 1–34, 2021.

\bibitem{retrieval:wei2019} J. Wei and K. Zou, ``Eda: Easy data augmentation techniques for boosting performance
on text classification tasks," \emph{arXiv:1901.11196}, 2019.

\bibitem{retrieval:karpathy2015} A. Karpathy and L. Fei-Fei,
``Deep visual-semantic alignments for generating image descriptions,"
in \emph{Proc. CVPR}, 2015.



\bibitem{retrieval:hendrycks2019} D. Hendrycks and T. Dietterich,
``Benchmarking neural network robustness to common corruptions and perturbations," in \emph{Proc. ICLR}, 2019.

\bibitem{retrieval:ren2015} S. Ren, K. He, R. Girshick, and J. Sun,
``Faster r-cnn:towards real-time object detection with region proposal networks," \emph{Advances in Neural Information Processing Systems (NIPS)}, 2015.


\bibitem{retrieval:jiang2019} H. Jiang, I. Misra, M. Rohrbach, E. Miller, and X. Chen,
``In defense of grid features for visual question answering,"
in \emph{Proc. ICCV}, 2019.

\bibitem{retrieval:devlin2019} J. Devlin and M. Chang and K. Lee and K. Toutanova,
``Bert: Pre-training of deep bidirectional transformers for language understanding,"
in \emph{arXiv:1810.04805}, 2019.

  
  










\end{thebibliography}
\end{document}